\documentclass[10pt,twocolumn,letterpaper]{article}

\usepackage{cvpr}              

\definecolor{cvprblue}{rgb}{0.21,0.49,0.74}

\usepackage{color,xcolor}
\usepackage{colortbl}
\usepackage{amssymb}
\usepackage{pifont}

\usepackage{tabu}
\usepackage{multirow}
\usepackage{amsmath,bm}

\usepackage{wrapfig}
\usepackage{adjustbox}

\usepackage{float}
\usepackage[accsupp]{axessibility}
\usepackage[pagebackref,breaklinks,colorlinks,allcolors=cvprblue]{hyperref}

\DeclareMathOperator{\jpeg}{JPEG}
\DeclareMathOperator{\mpeg}{MPEG}

\DeclareMathOperator{\dft}{DFT}

\DeclareMathOperator{\filter}{LP}

\DeclareMathOperator{\ddim}{DDIM}
\DeclareMathOperator{\norm}{Norm}

\def\paperID{4133} 
\def\confName{CVPR}
\def\confYear{2025}

\title{FADE: Frequency-Aware Diffusion Model Factorization for Video Editing}

\def\spaces{~~~~}

\author{Yixuan Zhu\textsuperscript{1}\spaces{}Haolin Wang\textsuperscript{1}\spaces{}Shilin Ma\textsuperscript{2}\spaces{}Wenliang Zhao\textsuperscript{1}\spaces{}Yansong Tang\textsuperscript{2}\spaces{}Lei Chen$^{1,\dagger}$\spaces{}Jie Zhou\textsuperscript{1}\\\\
\textsuperscript{1}Department of Automation, Tsinghua University \\
\textsuperscript{2}Tsinghua Shenzhen International Graduate School, Tsinghua University
}

\begin{document}

\twocolumn[{
    \renewcommand\twocolumn[1][]{#1}
    \maketitle
    \begin{center}
    \vspace{-15pt}
        \includegraphics[width=\linewidth]{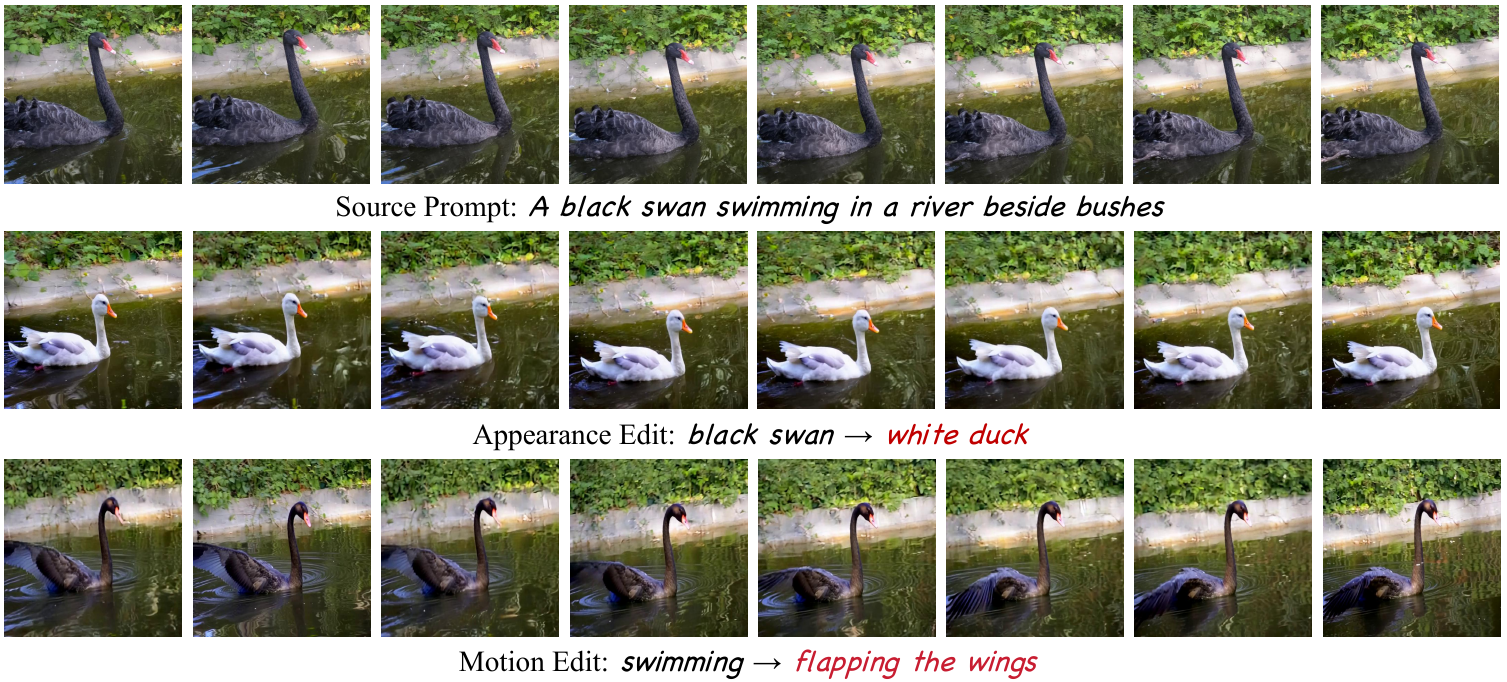}
        \captionof{figure}{\textbf{Diverse video editing results of \textit{FADE}.} Our training-free approach, utilizing frequency-aware factorization and modulation, achieves high-fidelity, coherent edits across a variety of video types. FADE handles both appearance and motion adjustments with impressive robustness, ensuring precise alignment with input prompts and maintaining temporal consistency. Best viewed in color.}
        \label{fig: teaser}
        \vspace{-5pt}
    \end{center}
}]
\renewcommand{\thefootnote}{}
\footnotetext{$\dagger$ Corresponding author}

\begin{abstract}
 Recent advancements in diffusion frameworks have significantly enhanced video editing, achieving high fidelity and strong alignment with textual prompts. However, conventional approaches using image diffusion models fall short in handling video dynamics, particularly for challenging temporal edits like motion adjustments. While current video diffusion models produce high-quality results, adapting them for efficient editing remains difficult due to the heavy computational demands that prevent the direct application of previous image editing techniques. To overcome these limitations, we introduce FADE—a training-free yet highly effective video editing approach that fully leverages the inherent priors from pre-trained video diffusion models via frequency-aware factorization. Rather than simply using these models, we first analyze the attention patterns within the video model to reveal how video priors are distributed across different components. Building on these insights, we propose a factorization strategy to optimize each component’s specialized role. Furthermore, we devise spectrum-guided modulation to refine the sampling trajectory with frequency domain cues, preventing information leakage and supporting efficient, versatile edits while preserving the basic spatial and temporal structure. Extensive experiments on real-world videos demonstrate that our method consistently delivers high-quality, realistic and temporally coherent editing results both qualitatively and quantitatively. Code is available at \url{https://github.com/EternalEvan/FADE}.
\end{abstract}    
\section{Introduction}
\label{sec:intro}
\begin{figure}[t]
    \centering
    \includegraphics[width=0.99\linewidth]{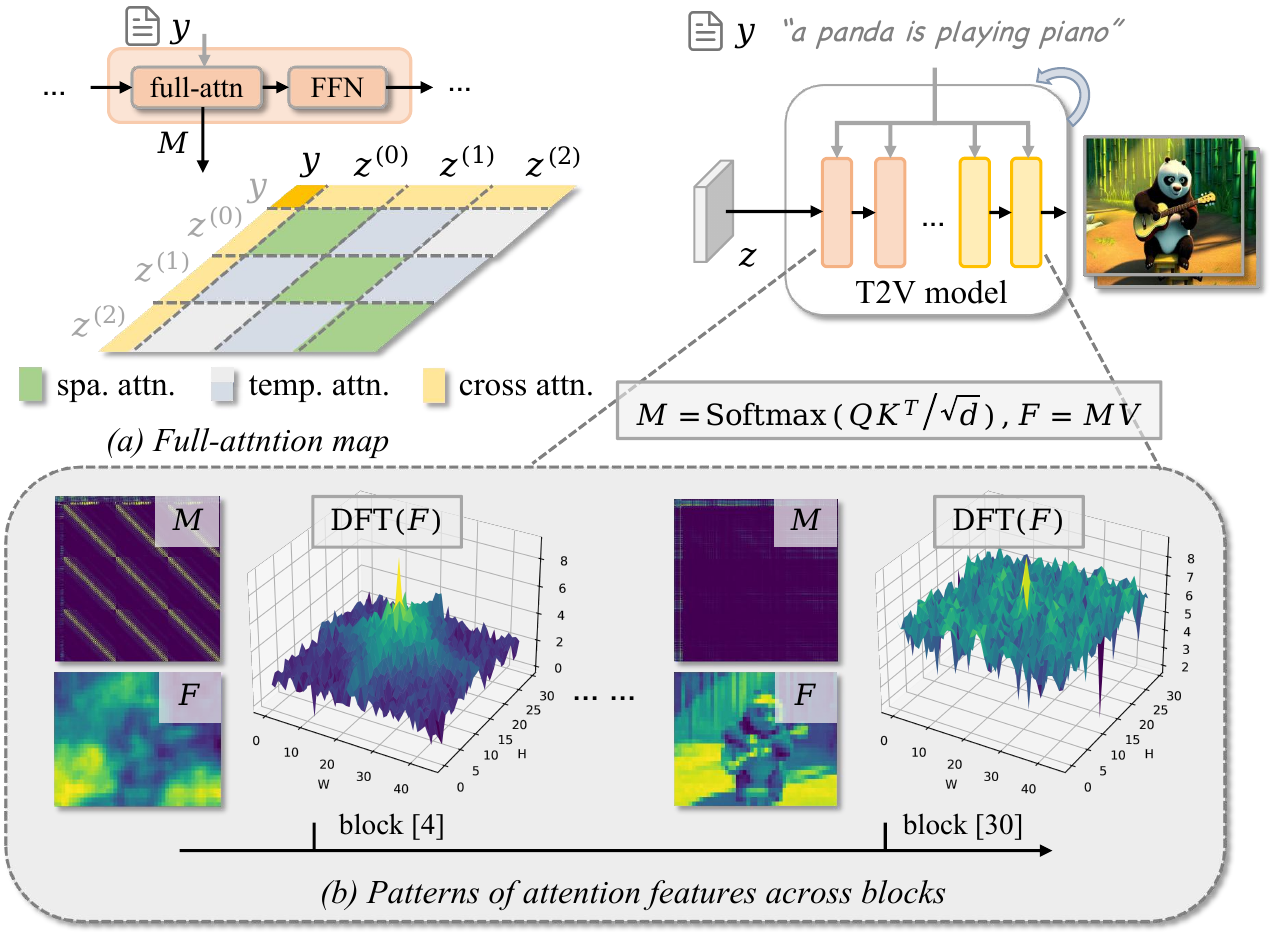}
    \caption{
    (a) In typical T2V models, video tokens flattened along the temporal axis yield full-attention maps split into cross, spatial, and temporal attention. (b) Video priors in T2V models are factorized based on attention feature patterns. We visualize the attention map $\bm M$ and result $\bm F$ along with its spectrum for two blocks. The early block sketches the foundational structure and movement, while the later block refines details. This insight allows us to leverage each block's unique role for more efficient editing.}
    \label{fig: attention}
\end{figure}

The objective of video editing is to modify specific visual content consistently across all frames while preserving the integrity of unchanged regions. Recent achievements in diffusion models have greatly elevated video editing capabilities~\cite{ho2022video-diffusion-models,ho2022imagen,villegas2022phenaki,singer2022make-a-video,molad2023dreamix,liu2024video-p2p,parmar2023pix2pix-zero}, bringing remarkable generative power to video editing tasks that require both spatial details and temporal coherence. While previous methods~\cite{hertz2022prompt2prompt,mokady2023null-text,kawar2023imagic,avrahami2022blended-edit,titov2024guide-and-rescale} have adapted text-to-image (T2I) diffusion models for prompt-driven image editing, their effectiveness diminishes when applied to video due to the complexities of managing spatial-temporal relationships across frames and a lack of generalized video prior knowledge. These limitations underscore the need for an efficient and robust editing strategy capable of handling the increased data and temporal coherence demands of videos. In response, we aim to design an accessible, high-quality editing solution that fully leverages the rich video priors embedded in pre-trained video diffusion models, enabling realistic, coherent, and versatile edits, including both appearance and motion adjustments, across complex video content.

The main challenge in video editing is balancing changes to the target object while preserving unedited content. Previous methods largely rely on the generative power of T2I models for controlled edits. For instance,~\cite{liu2024video-p2p,wang2023vid2vid-zero} achieve this through null-text inversion~\cite{mokady2023null-text}, which minimizes inversion errors via test-time optimization. However, this process requires extensive iterations to compute a series of null-text embeddings for every sampling step, making it time-consuming and limiting its real-world applicability. Another line of work is attention feature injection~\cite{qi2023fatezero,liu2024video-p2p,cao2023masactrl,shin2024edit-a-video}, where attention maps are modified by replacing or interleaving to maintain content consistency. Despite this, the absence of video priors often results in unrealistic temporal dynamics and limits motion edit capacity. Some methods, like~\cite{molad2023dreamix}, directly utilize video diffusion models by adding noise to the source video. This straightforward approach fails to fully exploit video priors, resulting in suboptimal editing quality. With the recent emergence of advanced text-to-video (T2V) models, new opportunities arise to harness their rich video knowledge. However, adapting previous T2I techniques to these models is challenging, given the extensive resources required, including memory, computation and time, to handle the increased data and complexity inherent in large-scale video networks. 

To overcome these challenges, we introduce FADE, a training-free yet potent approach for real-world video editing. Our goal is to harness the rich video priors about video appearance and dynamics within pre-trained T2V models to enable realistic and diverse editing. To achieve this, we start by examining the specific properties of these models. While previous work~\cite{ho2020ddpm} shows that diffusion models naturally emphasize different frequency components across sampling timesteps, we further observe that a similar frequency progression occurs block-wise within T2V models. We explore how video priors function in each block by analyzing their roles and patterns, revealing that early blocks sketch foundational spatial layouts and temporal dynamics while later blocks refine high-frequency details (Figure~\ref{fig: attention}). Leveraging this insight, we factorize the editing problem by the distinct functions of each component, enhancing editing efficiency. Then, instead of performing optimization or attention injection, we design a spectrum-guided modulation strategy that utilizes the full-attention outputs in these sketching blocks to derive frequency-aware guidance through 3D Fourier transformation. This modulation adjusts the sampling trajectory to preserve low-frequency structure from the source video while allowing high-frequency details to support flexible and intricate edits. This cue from the frequency domain not only mitigates information leakage during guidance computation but also unleashes the potential of video diffusion priors, enabling sophisticated spatial and temporal edits with high fidelity and coherence.

\begin{figure*}[htp]
    \centering
    \includegraphics[width=0.98\linewidth]{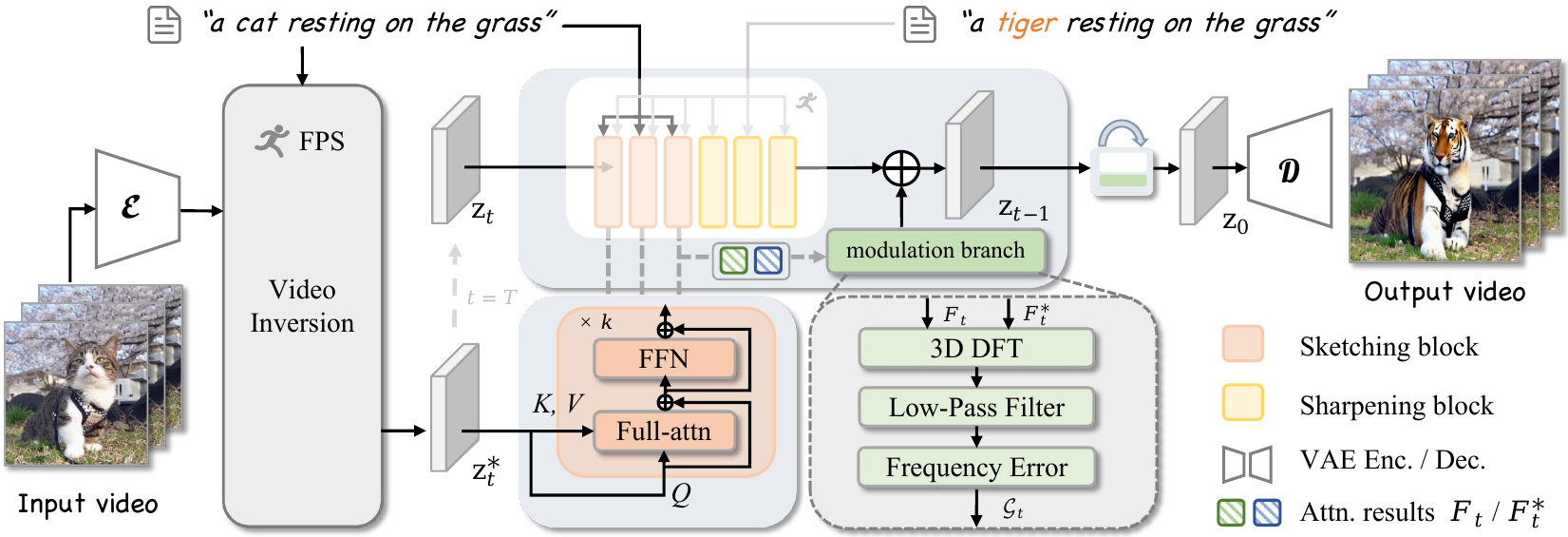}
    \caption{\textbf{The overall framework of \textit{FADE}.} FADE is a training-free framework for video editing. Given an input video, we first perform video inversion to obtain noise $\bm z_T$ and trajectory $\{\bm z_t^*\}_{t=0}^T$. Starting from this noise, we use sketching blocks in the T2V model to produce full-attention results $\bm F_t$ and $\bm F_t^*$ from $\bm z_t$ and $\bm z_t^*$. These results are transformed to the frequency domain, where a low-pass filter isolates spatial and temporal structures. The frequency error then modulates the sampling process, enabling high-quality, consistent edits.}
    \label{fig:pipeline}
    \vspace{-10pt}
\end{figure*}
We evaluate our method using extensive real-world video clips across various editing tasks. As shown in Figure~\ref{fig: teaser}, FADE exhibits proficiency in producing realistic and consistent edits, covering both appearance and motion modifications. Quantitative comparisons on video editing benchmark DAVIS~\cite{pont2017davis} demonstrate our approach achieves superior textual alignment and temporal consistency, with 0.3561 CLIP score and 43.57 OSV. Moreover, we maintain competitive input fidelity with 21.54 Mask-PSNR and 0.3297 LPIPS, preserving unedited regions effectively. Compared to previous methods, our use of general video priors with frequency-aware factorization delivers improved efficiency and a broader editing range, highlighting FADE's outstanding versatility and practicality.

\section{Related Works}
\noindent \textbf{Video Diffusion Models.} In recent years, denoising diffusion models~\cite{ho2020ddpm,song2020ddim} have emerged as a leading class of generative models, notable for their synthesis quality and controllability. After training on large-scale datasets, text-to-image (T2I) diffusion models~\cite{rombach2022high,dhariwal2021diffusion} exhibit exceptional image synthesis capability. For video generation, some works~\cite{blattmann2023stable-video-diffusion,ho2022vdm} adapt image diffusion models by inflating the denoising UNet~\cite{ronneberger2015unet} with temporal layers to handle video inputs. However, replicating the success of T2I models for video generation remains challenging due to the lack of large-scale video datasets and the increased data complexity. Most recently, Sora~\cite{brooks2024sora} and following studies~\cite{zheng1open-sora,yang2024cogvideox,ma2024latte} pioneer high-quality, high-resolution text-to-video (T2V) diffusion models. These models commonly use the diffusion transformer (DiT) architecture~\cite{peebles2023dit} to capture complex spatial and temporal coherence, outperforming previous UNet-based methods. However, this improved performance comes at the cost of high computational demands, mostly due to the extensive attention mechanism, which requires processing a large number of tokens. Limited research has explored the pattern within these models to reduce the computational intensity. To address this, we delve into their internal structure to analyze the roles of individual blocks, leveraging these insights to facilitate their applicability and efficiency in video editing.

\noindent \textbf{Video Editing.}
Significant progress in image editing~\cite{hertz2022prompt2prompt,mokady2023null-text,brooks2023instructpix2pix,meng2021sdedit,tumanyan2023plug-and-play,parmar2023pix2pix-zero} has driven advancements in video editing~\cite{liu2024video-p2p,geyer2023tokenflow,wu2023tune-a-video,molad2023dreamix,qi2023fatezero}. However, unique challenges in video editing arise from the need for consistent, coherent edits across frames, including both appearance and motion changes. Recent works~\cite{liu2024video-p2p,qi2023fatezero} adopt the attention feature injection techniques, manipulating various attention results to guide edits. They also employ model inflation to tune a T2I model with the input video following~\cite{wu2023tune-a-video}. These approaches, however, are inefficient since they require a separate model for each sample and lack the general video priors to handle motion edits. In contrast,~\cite{molad2023dreamix} utilizes a pre-trained video diffusion model for editing, beginning with corrupting the video by downsampling and noise. However, this corruption can result in global inconsistency with the source video and limits control over specific edits. Overall, there remains significant potential for advancing the use of video diffusion models in video editing. This application of the video diffusion model is nontrivial because (1) it requires maintaining global consistency in the generated video without relying on one-shot tuning or null-text optimization~\cite{mokady2023null-text}, and (2) applying previous attention feature injection methods is difficult with modern video models, as they involve dozens of transformer blocks with memory-intensive, full-attention computations. Our frequency-aware approach effectively addresses these challenges, setting our method apart from previous work.

\section{Method}
In this section, we present FADE, a training-free framework
that achieves real-world video editing with frequency-aware factorization of diffusion models. Our key idea is to investigate the inherent patterns in a pre-trained text-to-video diffusion model and activate each block's specific function for various video editing tasks. We will start by reviewing the background of video diffusion models, and then describe our designs
of FADE, including how to factorize video diffusion models according to frequency cues and adjust the sampling trajectory via spectrum-guided modulation. The overall framework of
our FADE is illustrated in Figure~\ref{fig:pipeline}.
\subsection{Preliminaries: Video Diffusion Models} 
Video diffusion models are a class of generative models that reconstruct the distribution of video data by modeling the reverse trajectory of a diffusion process. Given a random noise $\bm\epsilon\sim\mathcal{N}(0,\bm {I})$, the diffusion process is defined as:
\begin{equation}
    \bm z_t = \alpha_t \bm z_0 + \sigma_t \bm \epsilon,
    \label{eq: diffusion process}
\end{equation}
where $\bm z_0$ and $\bm z_t$ are the clean and noisy video, and $\{(\alpha_t, \sigma_t)\}^T_{t=1}$ denotes the noise schedule. With proper re-parameterization, the training objective of diffusion models can be expressed as:
\begin{equation}
    \mathcal{L}_{\rm DM}=\mathbb{E}_{\bm z_0,\bm \epsilon,t}\left[\Vert\bm\epsilon-\bm \epsilon_\theta(\bm z_t,t,\bm y)\Vert_2^2\right],
\end{equation}
where $\bm z_t$ is computed as Equation~\eqref{eq: diffusion process}. The denoising model $\bm \epsilon_\theta$ is learned to predict the noise given the prompt $\bm y$. Previous T2I models often implement $\bm \epsilon_\theta$ as UNet~\cite{ronneberger2015unet}. 
For video generation, recent models~\cite{brooks2024sora,ma2024latte,zheng1open-sora,yang2024cogvideox} usually employ a series of cascaded transformer blocks to capture the spatial-temporal relationships more effectively.

To generate a video in a small number of denoising steps, diffusion models usually use deterministic DDIM sampling:
\begin{equation}
    \bm z_{t-1} = \frac{\alpha_{t-1}}{\alpha_t}\bm z_t+(\sigma_{t-1}-\frac{\alpha_{t-1}}{\alpha_t}\sigma_t)\bm\epsilon_\theta(\bm z_t,t,\bm y).
    \label{eq: ddim}
\end{equation}
Conversely, to convert a video to its corresponding noise, we can apply DDIM inversion, the reverse of Equation~\eqref{eq: ddim}:
\begin{equation}
    \bm z_{t+1} = \frac{\alpha_{t+1}}{\alpha_t}\bm z_t+(\sigma_{t+1}-\frac{\alpha_{t+1}}{\alpha_t}\sigma_t)\bm\epsilon_\theta(\bm z_t,t,\bm y).
    \label{eq: ddim-inv}
\end{equation}
In practice, DDIM inversion introduces slight errors in every step, making it unreliable for fully reconstructing a real-world video or preserving fine details in the source video.   

\subsection{Frequency-Aware Factorization of T2V Model}
Modern T2V models typically employ cascades of transformer blocks as the denoising model, with each block containing attention layers that process extensive video tokens. Given that dozens of identical blocks are involved in calculating the same form of attention, a key question arises: \textit{Do all these blocks serve the same purpose and contribute equally to video synthesis?} To explore this, we analyze the behavior within each block of a pre-trained T2V model during video generation. Previous work~\cite{ho2020ddpm} shows that diffusion models naturally emphasize different frequency components across sampling timesteps. Inspired by this, we adopt a frequency-aware perspective to uncover if similar trends occur at the block level in video models. Specifically, we capture and visualize each block’s attention maps $\{\bm{M}_i\}_{i=1}^N$ and outputs $\{\bm{F}_i\}_{i=1}^N$, along with their Fourier spectra, over all $N$ blocks during sampling. As illustrated in Figure~\ref{fig: attention}, a typical attention map in T2V models contains three types of attention regions: cross-attention for multimodal alignments, spatial attention for capturing intra-frame spatial structures, and temporal attention for tracking dynamics across frames. Our observation is that spatial and temporal attention are densely aligned along diagonal lines in the early blocks, creating multiple parallel diagonals in the attention maps. The main diagonal line suggests that these early blocks focus on outlining the intra-frame spatial structures, defining key shapes and boundaries of the composition. The secondary diagonal lines indicate that these blocks are simultaneously establishing temporal correspondences across frames within the same regions. Collectively, these early blocks construct the low-frequency, foundational structure of the video, setting up basic object placement and movement. We refer to them as \textit{sketching blocks}. In contrast, the following blocks—which constitute the majority of the denoising model—display relatively sparse attention distribution, with scores more evenly spread across elements in the map. This implies that these later blocks focus on refining the high-frequency refinements, such as color, texture, and subtle vibration, providing nuanced adjustments for visual details. We call them \textit{sharpening blocks}, as their primary role is to enhance details, enriching the visual appeal of the final output.

The variation in attention outputs of these blocks further supports our inference. The initial sketching blocks produce blurry outputs, with spectra focused on low frequencies that capture basic spatial and temporal layouts. In contrast, the sharpening blocks reveal clear foreground and background regions, involving much more high-frequency details. This distinction suggests that these two classes of blocks process features of differing levels and serve different purposes. We further verify that each block’s function remains consistent throughout the sampling steps by analyzing these patterns across all steps. Thus, for video content editing—where maintaining underlying structures while adjusting finer details is essential—we can factorize each block's role, using sketching blocks for structural reconstruction and sharpening blocks for detailed generation. This division offers two key advantages: (1) it enables each block to perform optimally within its specialized function, and (2) it greatly reduces computational load by enabling frequency-aware guidance through selective use of blocks during sampling.
%
\subsection{Spectrum-Guided Modulation}
After factorizing the video model based on frequency-informed attention patterns, our next step is to activate each block's specific function and develop a robust sampling strategy that optimally balances reconstruction with generation. We begin by employing the inversion process described in Equation~\eqref{eq: ddim-inv} with a pre-trained video diffusion model $\bm\epsilon_\theta$. This transforms the real-world video $\bm z_0^*$ to an approximated noise $\bm z_T^*$ and produces a reverse trajectory $\{\bm z_t^*\}_{t=0}^T$. However, as discussed in prior works, DDIM inversion can introduce non-negligible errors, making it difficult to accurately reconstruct the source video from $\bm z_T^*$. This issue is particularly pronounced in video inversion, where the predicted noise $\bm z_T^*$ often deviates far from the standard Gaussian distribution. We, therefore, focus on refining the sampling process using the observed spectrum patterns to correct the inversion error and create an ideal trajectory $\{\bm z_t\}_{t=0}^T$ to a plausible edited video $\bm z_0$. Following previous image editing approaches~\cite{mokady2023null-text,titov2024guide-and-rescale}, we adopt a dual branch strategy that leverages the inversion trajectory to enhance consistency with the source video. However, existing attention feature injection methods require extensive memory due to full-attention maps and limit editing flexibility through restrictive attention fusion techniques. While these methods can preserve the spatial layout, they often struggle with more complex edits involving pose, shape and motion. 


To overcome these limitations, we propose a more flexible, spectrum-guided approach that factorizes visual information in the frequency domain to separate content for preservation and editing. This approach offers greater freedom for a wider range of edits. As shown in Figure~\ref{fig:pipeline}, our first design is to harness the sketching blocks, which care about the overall spatial layout and temporal movement. Rather than swapping or interleaving the attention maps directly, we gather the attention outputs of the source video $\bm z_t^*$ and the target video $\bm z_t$ within these blocks, denoted as $\bm{F}_t^*$ and $\bm{F}_t$, and use their difference to derive a guidance term for the sampling process. Specifically, these attention outputs are computed using the same source prompt $\bm y_{src}$ through the full-attention layers. This guidance term provides a more robust and adaptive way to adjust the sampling trajectory without directly altering the intermediate features within the diffusion model. Furthermore, since these features are derived from the sketching blocks, the guidance focuses on the coarse spatial and temporal content, such as basic layouts and object movements, leaving ample room for detailed edits to be made in the sharpening blocks. 
\begin{figure*}[htp]
    \centering
     \includegraphics[width=\linewidth]{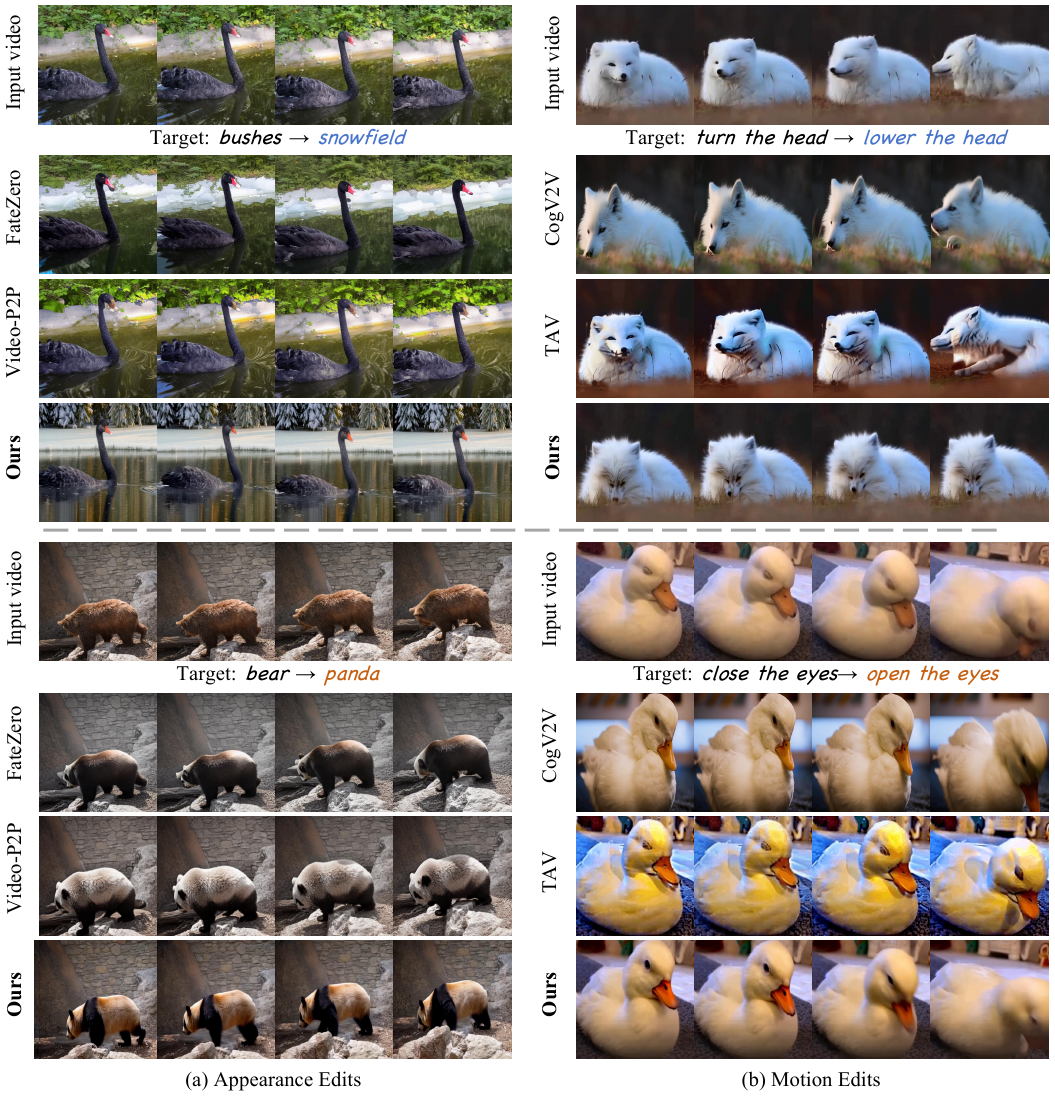}
     \setlength{\abovecaptionskip}{-5pt}
    \caption{\textbf{Qualitative comparisons on real-world videos.} We evaluate both (a) appearance and (b) motion edits. Our framework consistently achieves accurate textual alignment across frames and high fidelity in unedited regions. Notably, compared to other methods based on video models, we produce more coherent and realistic motion edits. More results are available in the supplementary materials.}
    \label{fig:main-exp}
\end{figure*}

However, native guidance can lead to information leakage from the source video to the target video, as the attention outputs contain the full range of visual content. To address this, it is crucial to determine which aspects of the content should be retained and which should be modified. For this purpose, we turn to the frequency domain for assistance. Real-world images and videos often exhibit redundancy across their spatial and temporal dimensions, meaning that most of their energy is concentrated in the low-frequency spectrum, while the high-frequency components contribute much less. This characteristic underpins popular compression algorithms such as $\jpeg$ for images and $\mpeg$ for videos, and we aim to apply this principle here. Since we cannot get a clean video during the denoising process, we use attention outputs composed of tokens representing the noisy video. Unlike the discrete cosine transform (DCT) used in the spatial domain for $\mpeg$, we employ the discrete Fourier transform on both spatial and temporal dimensions (3D DFT) to capture the spectral characteristics of intra-frame content and inter-frame motion:
\begin{equation}
    \mathcal{F}_t=\dft_{\rm 3D}(\bm{F}_t)\in\mathbb{R}^{h\times w \times \tau \times c},
\end{equation}
where $h$, $w$ and $\tau$ are the spatial and temporal size of the video tokens, and $c$ is the channel dimension. As we know, most of the energy in $\mathcal{F}_t$ is concentrated in the low-frequency range. Thus, with the observed attention patterns, we can assume that fundamental spatial structures and temporal movements are encapsulated within this range, while the high-frequency components contain Gaussian noise and specific details like texture and subtle shaking—elements we wish to suppress or remove during editing. To achieve this, we apply a low-pass filter to isolate the low-frequency content of $\mathcal{F}_t$. The same process is applied to $\mathcal{F}_t^*$ to compute the spectrum guidance $\mathcal{G}_t$:
\begin{equation}
    \mathcal{G}_t=\left\Vert \filter(\mathcal{F}_t)-\filter(\mathcal{F}_t^*)\right\Vert^2_2,
\end{equation}
where $\filter$ denotes the low-pass filter. This guidance measures the deviation of the target video $\bm z_t$ from the source video in terms of basic structure and movement. This distance is then used to modulate the trajectory during DDIM sampling by the gradient of $\mathcal{G}_t$ with respect to $\bm z_t$:
\begin{equation}
    \bm z_{t-1} = \ddim(\bm\epsilon_\theta,\bm z_t,t,\bm y_{tgt})-\lambda\norm(\nabla_{\bm z_t}\mathcal{G}_t),
\end{equation}
where $\ddim$ denotes the sampling process in Equation~\eqref{eq: ddim}, $\lambda$ is the guidance weight and $\norm$ normalizes the gradient to avoid numerical instability. This modulation strategy iteratively denoises the edited video towards the ideal direction, ensuring high consistency and realistic edits.

\subsection{Versatile Video Editing Framework}
We utilize a pre-trained T2V diffusion model for video inversion and sampling. This model leverages a 3D VAE, comprising an encoder $\mathcal{E}$ and a decoder $\mathcal{D}$, to enable spatial and temporal compression, facilitating a reversible transformation between pixel space and latent space. In addition to text embeddings, the T2V model incorporates FPS information to better handle video data. Unlike one-shot tuned video models from T2I models~\cite{liu2024video-p2p}, using a well-trained video diffusion model ensures a more stable understanding of the complex spatial and temporal relationships within videos. Moreover, this eliminates the need for model tuning, enhancing efficiency and reducing the necessity of storing individual models for specific videos. For edit types, we broadly categorize prompt-driven modifications into appearance edits and motion edits. Appearance edits involve altering nouns, such as changing the main subject or background, focusing on the texture, color, and style (\textit{e.g.}, transforming a ``cat" to a ``tiger" in Figure~\ref{fig:pipeline}). Motion edits, on the other hand, involve changing verbs, which can alter spatial and temporal dynamics. By applying spectrum-guided modulation to harness inherent video priors within the T2V model, our method achieves realistic motion edits, distinguishing it from most previous approaches.

\begin{table}
\centering
\caption{\textbf{Quantitative comparisons on real-world video data.} We compare our FADE framework to previous methods on both appearance and motion edits, measuring editing accuracy, input fidelity, temporal consistency, and human preference. FADE achieves competitive performance across diverse editing tasks.}
\adjustbox{width=\linewidth}{\begin{tabular}{lccccc}
\toprule
Method  &
CLIP $\uparrow$&
M.PSNR $\uparrow$ & 
LPIPS $\downarrow$  &
OSV $\downarrow$ &
PF $\uparrow$ \\
\midrule
\textit{appearance edits} \\
\midrule

Tune-A-Video~\cite{wu2023tune-a-video} &0.3522 &19.86 &0.4625 &35.01 & 0.12\\
Video-P2P~\cite{liu2024video-p2p} &0.3589 &20.57 &0.3199 &34.71 & 0.15\\
FateZero~\cite{qi2023fatezero} &0.3562 &20.65 &\textbf{0.3057} &33.23 & 0.29\\
CogVideoX-V2V~\cite{yang2024cogvideox}  &0.3754 & 18.96 &0.4811 &31.45 & 0.09\\
\midrule
\rowcolor{gray!25}
FADE (Ours) 
 & \textbf{0.3762}& \textbf{20.69}
 &0.3085& \textbf{31.36} & \textbf{0.35} \\
 
\midrule
\textit{motion edits} \\
\midrule

Tune-A-Video~\cite{wu2023tune-a-video} &0.3281	&18.68 &	0.4637&	35.85&	0.10\\
Video-P2P~\cite{liu2024video-p2p} & 0.3314	&18.93	&0.3885	&35.44	&0.15\\
FateZero~\cite{qi2023fatezero} & 0.3259	&19.02	&0.3712	&34.47	&0.13\\
CogVideoX-V2V~\cite{yang2024cogvideox}  &0.3678	&18.17	&0.4928	&35.52	&0.19\\
\midrule
\rowcolor{gray!25}
FADE (Ours) 
 & \textbf{0.3683}	&\textbf{19.26}	&\textbf{0.3692}	&\textbf{32.28}	&\textbf{0.43} \\

\bottomrule
\end{tabular}
}

\label{table:quant-cmp}
\vspace{-15pt}
\end{table}
\section{Experiments}
To verify the effectiveness of our method, we conduct comprehensive experiments across various video editing tasks. We will start by describing the experimental settings and then present our main results. We will also provide detailed ablation studies and analyses of our approach to highlight the contributions of each component in our framework and to validate the impact of our design choices. 
\subsection{Experiment Setups}
\noindent \textbf{Implementation Details.} To utilize video generation priors, we adopt a recent, widely used T2V model~\cite{yang2024cogvideox} for video inversion and generation. This model's denoising component employs the DiT architecture with 48 transformer blocks, each containing a full-attention layer. Based on observed attention map patterns, we designate the first 4 blocks as sketching blocks. We calculate the guidance over the interval $[0,0.6T]$ of the DDIM sampling steps, with total steps set to $T=50$. To generate textual prompts from the source video, we use a multimodal language model~\cite{li2022blip} as the captioner. The guidance weight varies between 10 and 15, tailored to each editing task. We perform all evaluations on a single NVIDIA LS20 GPU with 48GB VRAM.

\noindent\textbf{Datasets and Metrics.}
We evaluate our method using the DAVIS dataset~\cite{pont2017davis} as well as various in-the-wild videos, following the approach of previous studies~\cite{liu2024video-p2p,qi2023fatezero}. Alongside qualitative results, we provide quantitative evaluations across multiple aspects using different metrics. Editing accuracy is assessed through CLIP score~\cite{radford2021clip}, which reflects the degree of textual alignment. To evaluate temporal consistency, a key factor for video coherence, we compute object semantic variance (OSV), following~\cite{liu2024video-p2p}, to capture semantic consistency across frames. Additionally, we assess fidelity in unedited regions by calculating Mask PSNR and LPIPS~\cite{zhang2018lpips}, which quantify the preservation of content that should remain unchanged. Furthermore, given the importance of visual quality in video editing, we conduct a user study to gauge human preference across all editing tasks.

\subsection{Main Results}
We compare our editing method with state-of-the-art approaches, including Video-P2P~\cite{liu2024video-p2p}, FateZero~\cite{qi2023fatezero}, Tune-A-Video (TAV)~\cite{wu2023tune-a-video} and CogVideoX-V2V (CogV2V)~\cite{yang2024cogvideox}. As illustrated in Figure~\ref{fig:main-exp}, our method consistently yields superior results across all aspects when compared to previous methods. The editing results of Video-P2P and FateZero exhibit significant limitations, often failing to modify content accurately, resulting in noticeable textual mismatches. Similarly, while TAV and CogV2V leverage T2V models, they struggle with motion editing, producing subpar results in capturing dynamic changes. Leveraging the inherent priors from the T2V model and spectrum-guided modulation, our method achieves the best performance in preserving local structure, maintaining temporal consistency, and ensuring text-video alignment. We evaluate our method on 20 videos and measure 5 metrics for quantitative analysis. As shown in Table~\ref{table:quant-cmp}, our method achieves the highest CLIP and OSV scores, demonstrating its superiority in textual alignment and temporal consistency. We also obtain competitive Mask PSNR and LPIPS, indicating that our method preserves unedited regions effectively. Moreover, we achieve the highest human preference score, reflecting the favorable visual quality and appeal of our edits across tasks. Notably, our method operates without any optimization or fine-tuning, completing various edits on real-world videos in approximately 3 minutes, a significant efficiency gain compared to previous methods, which require over 15 minutes.

\begin{figure}[t]
    \centering
    \includegraphics[width=0.99\linewidth]{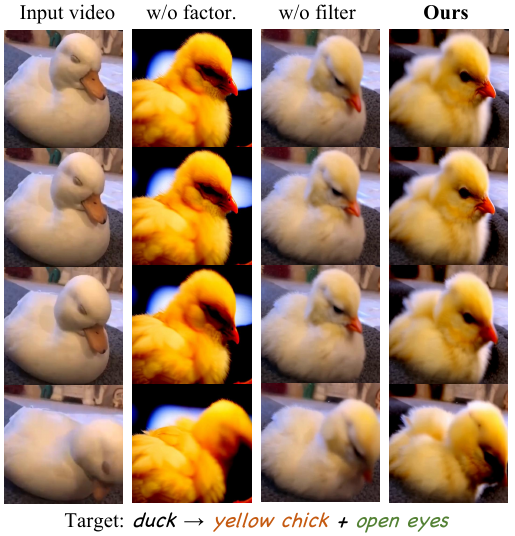}
    \caption{
    \textbf{Qualitative comparisons of the ablations.} We find the variant frameworks fall short in terms of editing accuracy, temporal consistency and input fidelity, yielding unsatisfactory results.}
    \label{fig: ablation}
    \vspace{-15pt}
\end{figure}
\subsection{Analysis}
\noindent \textbf{Efficient Editing via Sketching Blocks.} We assess the function of different blocks in the T2V model by testing various block combinations to calculate modulation, holding other settings constant. We consider three configurations: (1) using sketching blocks only (ours), (2) symmetrically combining sketching and sharpening blocks (symm. blocks), and (3) using all blocks without factorization (w/o factor.). As shown in Table~\ref{table: ablations}, configuration (1) provides comparable results to configurations (2) and (3), despite requiring fewer blocks for modulation. Although configurations (2) and (3) offer slightly improved fidelity, they significantly increase test time and memory demands. Figure~\ref{fig: ablation} further demonstrates that relying on only the first four transformer blocks, or sketching blocks, achieves the best editing performance. Interestingly, contrary to common assumptions, including the sharpening blocks actually diminishes editing quality. We assume this is because the sharpening blocks primarily focus on elements like color, texture, and subtle vibrations, which often do not need to remain unchanged in most video editing tasks. Relying on these blocks for guidance may, therefore, misdirect the model, diverting its attention from building the foundational layout and movement of the input video, ultimately resulting in reduced performance. These results underscore the effectiveness of our frequency-aware factorization approach, demonstrating that our selective focus on sketching blocks provides an optimal balance between maintaining high-quality edits and achieving computational efficiency.

\noindent \textbf{Impact of Spectrum-Guided Modulation.} In Figure~\ref{fig: ablation} and Table~\ref{table: ablations}, we examine our spectrum-guided modulation by removing the low-pass filter (w/o filter) and analyzing its effect on the editing performance. When we use the entire spectrum of attention outputs, high-frequency components are retained in the guidance term, causing the target object to retain too much of the source’s original characteristics, which weakens the text-video alignment. Experiments show that using about two-thirds of the frequency components in the guidance term balances reconstruction and editing quality. The quantitative results further underscore the efficacy of spectrum-guided modulation, as shown by superior performance on key metrics like textual alignment and temporal consistency. By selectively modulating the frequency components, our method achieves precise edits that align well with the given textual prompt and maintain consistency across frames, both of which are crucial for high-quality video editing. These findings highlight that spectrum-guided modulation not only supports a stable structural foundation for the edits but also enhances flexibility, allowing detailed adjustments for various editing tasks while ensuring input fidelity across the video.

\noindent \textbf{Effective Exploitation of Video Diffusion Model.} Our framework capitalizes on the rich video priors embedded in the pre-trained T2V model, enabling more effective utilization of the model’s inherent knowledge of spatial layouts, temporal coherence, and object motions. The video diffusion model captures nuanced understanding across frames, making it an ideal foundation for edits that require both high fidelity and temporal consistency. To validate our superior exploitation of these priors, we compare FADE with CogVideoX-V2V~\cite{yang2024cogvideox}, which uses the same underlying T2V model. As shown in Figure~\ref{fig:main-exp} and Table~\ref{table:quant-cmp}, FADE outperforms it both qualitatively and quantitatively across various editing tasks, demonstrating our framework’s ability to harness the model’s rich knowledge more effectively. 

\begin{table}
\centering
\caption{\textbf{Ablations studies.} We perform ablations to verify the effectiveness of the components in FADE and the impact of the model factorization. We find that FADE achieves similar performance with only a subset of sketching blocks, compared to using symmetric or all blocks. Moreover, spectrum-guided modulation further enhances both preservation and precision in edits.}
\adjustbox{width=\linewidth}{\begin{tabular}{lccccc}
\toprule
Method  &
CLIP $\uparrow$&
M.PSNR $\uparrow$ & 
LPIPS $\downarrow$  &
OSV $\downarrow$ & Test-time\\
\midrule

symm. blocks  &0.3659 &20.73 &0.3367& 32.61 & 5 min\\

w/o factor. &0.3691 &\textbf{20.94} &\textbf{0.3328} & 32.05& 12 min\\

w/o filter &0.3612 &20.89 &0.3364 & 32.28& 3 min\\

\midrule
\rowcolor{gray!25}
FADE (Ours) 
 & \textbf{0.3728}& 20.87
 &0.3352 & \textbf{31.77}  & 3 min\\

\bottomrule
\end{tabular}
}

\label{table: ablations}
\vspace{-10pt}
\end{table}

\noindent \textbf{Limitations.} Despite FADE’s effectiveness, its performance relies on the underlying video model and struggles with heavy occlusions requiring advanced temporal reasoning. We will address these in future work.

\section{Conclusion}
In this paper, we present FADE, a novel, training-free framework tailored for a diverse range of video editing tasks, encompassing both appearance and motion adjustments. Through frequency-aware factorization of the T2V model, FADE efficiently harnesses the model’s inherent structural and temporal patterns, allowing each block to excel in its designated role. We further design the spectrum-guided modulation, which adjusts the sampling trajectory using frequency cues to preserve foundational video elements and enhance text-video alignment. Extensive experiments demonstrate our framework's competitive performance and generalization across various video editing challenges. We envision our work will inspire future research on versatile video editing and efficient video diffusion model.
\noindent
\textbf{Acknowledgement.} This work was supported in part by the National Key Research and Development Program of China under Grant 2024YFB4708100, and in part by the National Natural Science Foundation of China under Grant Grant 62321005, Grant 62306031

{
    \small
    \bibliographystyle{ieeenat_fullname}
    \bibliography{main}

\begin{thebibliography}{38}
\providecommand{\natexlab}[1]{#1}
\providecommand{\url}[1]{\texttt{#1}}
\expandafter\ifx\csname urlstyle\endcsname\relax
  \providecommand{\doi}[1]{doi: #1}\else
  \providecommand{\doi}{doi: \begingroup \urlstyle{rm}\Url}\fi

\bibitem[Avrahami et~al.(2022)Avrahami, Lischinski, and Fried]{avrahami2022blended-edit}
Omri Avrahami, Dani Lischinski, and Ohad Fried.
\newblock Blended diffusion for text-driven editing of natural images.
\newblock In \emph{CVPR}, pages 18208--18218, 2022.

\bibitem[Blattmann et~al.(2023)Blattmann, Dockhorn, Kulal, Mendelevitch, Kilian, Lorenz, Levi, English, Voleti, Letts, et~al.]{blattmann2023stable-video-diffusion}
Andreas Blattmann, Tim Dockhorn, Sumith Kulal, Daniel Mendelevitch, Maciej Kilian, Dominik Lorenz, Yam Levi, Zion English, Vikram Voleti, Adam Letts, et~al.
\newblock Stable video diffusion: Scaling latent video diffusion models to large datasets.
\newblock \emph{arXiv preprint arXiv:2311.15127}, 2023.

\bibitem[Brooks et~al.(2023)Brooks, Holynski, and Efros]{brooks2023instructpix2pix}
Tim Brooks, Aleksander Holynski, and Alexei~A Efros.
\newblock Instructpix2pix: Learning to follow image editing instructions.
\newblock In \emph{CVPR}, pages 18392--18402, 2023.

\bibitem[Brooks et~al.(2024)Brooks, Peebles, Holmes, DePue, Guo, Jing, Schnurr, Taylor, Luhman, Luhman, et~al.]{brooks2024sora}
Tim Brooks, Bill Peebles, Connor Holmes, Will DePue, Yufei Guo, Li Jing, David Schnurr, Joe Taylor, Troy Luhman, Eric Luhman, et~al.
\newblock Video generation models as world simulators.
\newblock \emph{URL https://openai. com/research/video-generation-models-as-world-simulators}, 3, 2024.

\bibitem[Cao et~al.(2023)Cao, Wang, Qi, Shan, Qie, and Zheng]{cao2023masactrl}
Mingdeng Cao, Xintao Wang, Zhongang Qi, Ying Shan, Xiaohu Qie, and Yinqiang Zheng.
\newblock Masactrl: Tuning-free mutual self-attention control for consistent image synthesis and editing.
\newblock In \emph{ICCV}, pages 22560--22570, 2023.

\bibitem[Dhariwal and Nichol(2021)]{dhariwal2021diffusion}
Prafulla Dhariwal and Alexander Nichol.
\newblock Diffusion models beat gans on image synthesis.
\newblock \emph{NeurIPS}, 34:\penalty0 8780--8794, 2021.

\bibitem[Geyer et~al.(2023)Geyer, Bar-Tal, Bagon, and Dekel]{geyer2023tokenflow}
Michal Geyer, Omer Bar-Tal, Shai Bagon, and Tali Dekel.
\newblock Tokenflow: Consistent diffusion features for consistent video editing.
\newblock \emph{arXiv preprint arXiv:2307.10373}, 2023.

\bibitem[Hertz et~al.(2022)Hertz, Mokady, Tenenbaum, Aberman, Pritch, and Cohen-Or]{hertz2022prompt2prompt}
Amir Hertz, Ron Mokady, Jay Tenenbaum, Kfir Aberman, Yael Pritch, and Daniel Cohen-Or.
\newblock Prompt-to-prompt image editing with cross attention control.
\newblock \emph{arXiv preprint arXiv:2208.01626}, 2022.

\bibitem[Ho et~al.(2020)Ho, Jain, and Abbeel]{ho2020ddpm}
Jonathan Ho, Ajay Jain, and Pieter Abbeel.
\newblock Denoising diffusion probabilistic models.
\newblock \emph{NeurIPS}, 33:\penalty0 6840--6851, 2020.

\bibitem[Ho et~al.(2022{\natexlab{a}})Ho, Chan, Saharia, Whang, Gao, Gritsenko, Kingma, Poole, Norouzi, Fleet, et~al.]{ho2022imagen}
Jonathan Ho, William Chan, Chitwan Saharia, Jay Whang, Ruiqi Gao, Alexey Gritsenko, Diederik~P Kingma, Ben Poole, Mohammad Norouzi, David~J Fleet, et~al.
\newblock Imagen video: High definition video generation with diffusion models.
\newblock \emph{arXiv preprint arXiv:2210.02303}, 2022{\natexlab{a}}.

\bibitem[Ho et~al.(2022{\natexlab{b}})Ho, Salimans, Gritsenko, Chan, Norouzi, and Fleet]{ho2022vdm}
Jonathan Ho, Tim Salimans, Alexey Gritsenko, William Chan, Mohammad Norouzi, and David~J Fleet.
\newblock Video diffusion models.
\newblock \emph{NeurIPS}, 35:\penalty0 8633--8646, 2022{\natexlab{b}}.

\bibitem[Ho et~al.(2022{\natexlab{c}})Ho, Salimans, Gritsenko, Chan, Norouzi, and Fleet]{ho2022video-diffusion-models}
Jonathan Ho, Tim Salimans, Alexey Gritsenko, William Chan, Mohammad Norouzi, and David~J Fleet.
\newblock Video diffusion models.
\newblock \emph{NeurIPS}, 35:\penalty0 8633--8646, 2022{\natexlab{c}}.

\bibitem[Kawar et~al.(2023)Kawar, Zada, Lang, Tov, Chang, Dekel, Mosseri, and Irani]{kawar2023imagic}
Bahjat Kawar, Shiran Zada, Oran Lang, Omer Tov, Huiwen Chang, Tali Dekel, Inbar Mosseri, and Michal Irani.
\newblock Imagic: Text-based real image editing with diffusion models.
\newblock In \emph{CVPR}, pages 6007--6017, 2023.

\bibitem[Li et~al.(2022)Li, Li, Xiong, and Hoi]{li2022blip}
Junnan Li, Dongxu Li, Caiming Xiong, and Steven Hoi.
\newblock Blip: Bootstrapping language-image pre-training for unified vision-language understanding and generation.
\newblock In \emph{ICML}, pages 12888--12900. PMLR, 2022.

\bibitem[Liu et~al.(2024)Liu, Zhang, Li, Lin, and Jia]{liu2024video-p2p}
Shaoteng Liu, Yuechen Zhang, Wenbo Li, Zhe Lin, and Jiaya Jia.
\newblock Video-p2p: Video editing with cross-attention control.
\newblock In \emph{CVPR}, pages 8599--8608, 2024.

\bibitem[Ma et~al.(2024)Ma, Wang, Jia, Chen, Liu, Li, Chen, and Qiao]{ma2024latte}
Xin Ma, Yaohui Wang, Gengyun Jia, Xinyuan Chen, Ziwei Liu, Yuan-Fang Li, Cunjian Chen, and Yu Qiao.
\newblock Latte: Latent diffusion transformer for video generation.
\newblock \emph{arXiv preprint arXiv:2401.03048}, 2024.

\bibitem[Meng et~al.(2021)Meng, He, Song, Song, Wu, Zhu, and Ermon]{meng2021sdedit}
Chenlin Meng, Yutong He, Yang Song, Jiaming Song, Jiajun Wu, Jun-Yan Zhu, and Stefano Ermon.
\newblock Sdedit: Guided image synthesis and editing with stochastic differential equations.
\newblock In \emph{ICLR}, 2021.

\bibitem[Mokady et~al.(2023)Mokady, Hertz, Aberman, Pritch, and Cohen-Or]{mokady2023null-text}
Ron Mokady, Amir Hertz, Kfir Aberman, Yael Pritch, and Daniel Cohen-Or.
\newblock Null-text inversion for editing real images using guided diffusion models.
\newblock In \emph{CVPR}, pages 6038--6047, 2023.

\bibitem[Molad et~al.(2023)Molad, Horwitz, Valevski, Acha, Matias, Pritch, Leviathan, and Hoshen]{molad2023dreamix}
Eyal Molad, Eliahu Horwitz, Dani Valevski, Alex~Rav Acha, Yossi Matias, Yael Pritch, Yaniv Leviathan, and Yedid Hoshen.
\newblock Dreamix: Video diffusion models are general video editors.
\newblock \emph{arXiv preprint arXiv:2302.01329}, 2023.

\bibitem[Parmar et~al.(2023)Parmar, Kumar~Singh, Zhang, Li, Lu, and Zhu]{parmar2023pix2pix-zero}
Gaurav Parmar, Krishna Kumar~Singh, Richard Zhang, Yijun Li, Jingwan Lu, and Jun-Yan Zhu.
\newblock Zero-shot image-to-image translation.
\newblock In \emph{ACM SIGGRAPH}, pages 1--11, 2023.

\bibitem[Peebles and Xie(2023)]{peebles2023dit}
William Peebles and Saining Xie.
\newblock Scalable diffusion models with transformers.
\newblock In \emph{ICCV}, pages 4195--4205, 2023.

\bibitem[Pont-Tuset et~al.(2017)Pont-Tuset, Perazzi, Caelles, Arbel{\'a}ez, Sorkine-Hornung, and Van~Gool]{pont2017davis}
Jordi Pont-Tuset, Federico Perazzi, Sergi Caelles, Pablo Arbel{\'a}ez, Alex Sorkine-Hornung, and Luc Van~Gool.
\newblock The 2017 davis challenge on video object segmentation.
\newblock \emph{arXiv preprint arXiv:1704.00675}, 2017.

\bibitem[Qi et~al.(2023)Qi, Cun, Zhang, Lei, Wang, Shan, and Chen]{qi2023fatezero}
Chenyang Qi, Xiaodong Cun, Yong Zhang, Chenyang Lei, Xintao Wang, Ying Shan, and Qifeng Chen.
\newblock Fatezero: Fusing attentions for zero-shot text-based video editing.
\newblock In \emph{ICCV}, pages 15932--15942, 2023.

\bibitem[Radford et~al.(2021)Radford, Kim, Hallacy, Ramesh, Goh, Agarwal, Sastry, Askell, Mishkin, Clark, et~al.]{radford2021clip}
Alec Radford, Jong~Wook Kim, Chris Hallacy, Aditya Ramesh, Gabriel Goh, Sandhini Agarwal, Girish Sastry, Amanda Askell, Pamela Mishkin, Jack Clark, et~al.
\newblock Learning transferable visual models from natural language supervision.
\newblock In \emph{ICML}, pages 8748--8763. PMLR, 2021.

\bibitem[Rombach et~al.(2022)Rombach, Blattmann, Lorenz, Esser, and Ommer]{rombach2022high}
Robin Rombach, Andreas Blattmann, Dominik Lorenz, Patrick Esser, and Bj{\"o}rn Ommer.
\newblock High-resolution image synthesis with latent diffusion models.
\newblock In \emph{CVPR}, pages 10684--10695, 2022.

\bibitem[Ronneberger et~al.(2015)Ronneberger, Fischer, and Brox]{ronneberger2015unet}
Olaf Ronneberger, Philipp Fischer, and Thomas Brox.
\newblock U-net: Convolutional networks for biomedical image segmentation.
\newblock In \emph{MICCAI}, pages 234--241. Springer, 2015.

\bibitem[Shin et~al.(2024)Shin, Kim, Lee, Lee, and Yoon]{shin2024edit-a-video}
Chaehun Shin, Heeseung Kim, Che~Hyun Lee, Sang-gil Lee, and Sungroh Yoon.
\newblock Edit-a-video: Single video editing with object-aware consistency.
\newblock In \emph{ACML}, pages 1215--1230. PMLR, 2024.

\bibitem[Singer et~al.(2022)Singer, Polyak, Hayes, Yin, An, Zhang, Hu, Yang, Ashual, Gafni, et~al.]{singer2022make-a-video}
Uriel Singer, Adam Polyak, Thomas Hayes, Xi Yin, Jie An, Songyang Zhang, Qiyuan Hu, Harry Yang, Oron Ashual, Oran Gafni, et~al.
\newblock Make-a-video: Text-to-video generation without text-video data.
\newblock \emph{arXiv preprint arXiv:2209.14792}, 2022.

\bibitem[Song et~al.(2020)Song, Meng, and Ermon]{song2020ddim}
Jiaming Song, Chenlin Meng, and Stefano Ermon.
\newblock Denoising diffusion implicit models.
\newblock In \emph{ICLR}, 2020.

\bibitem[Titov et~al.(2024)Titov, Khalmatova, Ivanova, Vetrov, and Alanov]{titov2024guide-and-rescale}
Vadim Titov, Madina Khalmatova, Alexandra Ivanova, Dmitry Vetrov, and Aibek Alanov.
\newblock Guide-and-rescale: Self-guidance mechanism for effective tuning-free real image editing.
\newblock \emph{arXiv preprint arXiv:2409.01322}, 2024.

\bibitem[Tumanyan et~al.(2023)Tumanyan, Geyer, Bagon, and Dekel]{tumanyan2023plug-and-play}
Narek Tumanyan, Michal Geyer, Shai Bagon, and Tali Dekel.
\newblock Plug-and-play diffusion features for text-driven image-to-image translation.
\newblock In \emph{CVPR}, pages 1921--1930, 2023.

\bibitem[Villegas et~al.(2022)Villegas, Babaeizadeh, Kindermans, Moraldo, Zhang, Saffar, Castro, Kunze, and Erhan]{villegas2022phenaki}
Ruben Villegas, Mohammad Babaeizadeh, Pieter-Jan Kindermans, Hernan Moraldo, Han Zhang, Mohammad~Taghi Saffar, Santiago Castro, Julius Kunze, and Dumitru Erhan.
\newblock Phenaki: Variable length video generation from open domain textual descriptions.
\newblock In \emph{ICLR}, 2022.

\bibitem[Wang et~al.(2023)Wang, Jiang, Xie, Liu, Chen, Cao, Wang, and Shen]{wang2023vid2vid-zero}
Wen Wang, Yan Jiang, Kangyang Xie, Zide Liu, Hao Chen, Yue Cao, Xinlong Wang, and Chunhua Shen.
\newblock Zero-shot video editing using off-the-shelf image diffusion models.
\newblock \emph{arXiv preprint arXiv:2303.17599}, 2023.

\bibitem[Wu et~al.(2023)Wu, Ge, Wang, Lei, Gu, Shi, Hsu, Shan, Qie, and Shou]{wu2023tune-a-video}
Jay~Zhangjie Wu, Yixiao Ge, Xintao Wang, Stan~Weixian Lei, Yuchao Gu, Yufei Shi, Wynne Hsu, Ying Shan, Xiaohu Qie, and Mike~Zheng Shou.
\newblock Tune-a-video: One-shot tuning of image diffusion models for text-to-video generation.
\newblock In \emph{ICCV}, pages 7623--7633, 2023.

\bibitem[Yang et~al.(2024)Yang, Teng, Zheng, Ding, Huang, Xu, Yang, Hong, Zhang, Feng, et~al.]{yang2024cogvideox}
Zhuoyi Yang, Jiayan Teng, Wendi Zheng, Ming Ding, Shiyu Huang, Jiazheng Xu, Yuanming Yang, Wenyi Hong, Xiaohan Zhang, Guanyu Feng, et~al.
\newblock Cogvideox: Text-to-video diffusion models with an expert transformer.
\newblock \emph{arXiv preprint arXiv:2408.06072}, 2024.

\bibitem[Zhang et~al.(2018)Zhang, Isola, Efros, Shechtman, and Wang]{zhang2018lpips}
Richard Zhang, Phillip Isola, Alexei~A Efros, Eli Shechtman, and Oliver Wang.
\newblock The unreasonable effectiveness of deep features as a perceptual metric.
\newblock In \emph{CVPR}, pages 586--595, 2018.

\bibitem[Zheng et~al.(2024)Zheng, Peng, Yang, Shen, Li, Liu, Zhou, Li, and You]{zheng1open-sora}
Zangwei Zheng, Xiangyu Peng, Tianji Yang, Chenhui Shen, Shenggui Li, Hongxin Liu, Yukun Zhou, Tianyi Li, and Yang You.
\newblock Open-sora: Democratizing efficient video production for all.
\newblock \emph{URL https://github. com/hpcaitech/Open-Sora}, 2024.

\bibitem[Zhou et~al.(2022)Zhou, Girdhar, Joulin, Kr{\"a}henb{\"u}hl, and Misra]{zhou2022detecting}
Xingyi Zhou, Rohit Girdhar, Armand Joulin, Philipp Kr{\"a}henb{\"u}hl, and Ishan Misra.
\newblock Detecting twenty-thousand classes using image-level supervision.
\newblock In \emph{ECCV}, 2022.

\end{thebibliography}
}

\clearpage
\setcounter{page}{1}
\renewcommand\thesection{\Alph{section}} 
\renewcommand\thetable{\Alph{table}}
\renewcommand\thefigure{\Alph{figure}}
\setcounter{section}{0}
\setcounter{figure}{0}
\setcounter{table}{0}

\twocolumn[{
    \renewcommand\twocolumn[1][]{#1}
    \maketitlesupplementary
    \begin{center}
        \includegraphics[width=\linewidth]{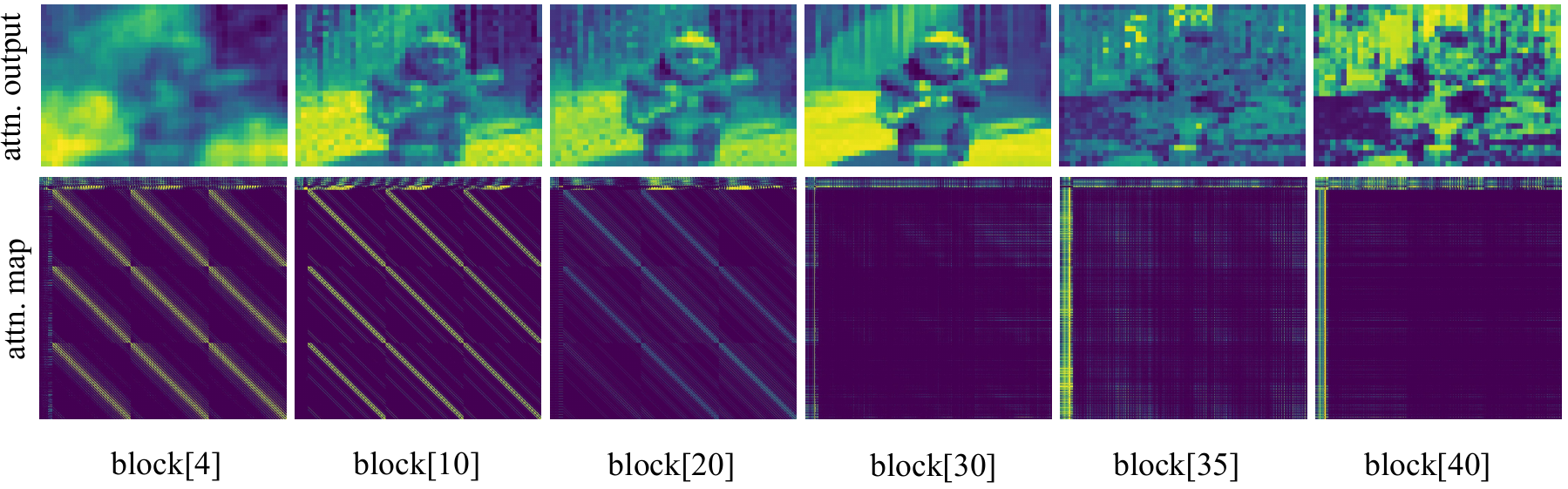}
        \captionof{figure}{\textbf{Visualization of attention outputs and attention maps.} We can clearly observe that the patterns of the attention features across blocks. Early blocks primarily process low-frequency structures, reflected in their blurred attention outputs and maps with multiple diagonal patterns. Conversely, later blocks focus on high-frequency details, producing attention maps with a uniformly distributed pattern.}
        \label{fig: attn-supp}
    \end{center}
}]

\appendix


\section{Detailed Implementations}
\label{sec:a}
To leverage video generation priors, we employ the widely used T2V model, CogVideoX-5b~\cite{yang2024cogvideox}, for video inversion and generation. In the inversion stage, the resolution of the input video is first adjusted to $720 \times 480$ to comply with the resolution constraint of CogVideoX-5b. Additionally, since the model is trained with long prompts, a chatbot is utilized to enrich the input prompt with additional details. Subsequently, DDIM inversion is performed with the total number of steps set to $T=50$, yielding the latent trajectories. In the editing stage, the first four blocks are designated as sketching blocks. To capture spectral characteristics, we apply 3D DFT on the attention output of these blocks. Additionally, the output of the last block in DiT is utilized to compute an auxiliary guidance term. The guidance mechanism is applied during the interval $[0, 0.6T]$ of the DDIM sampling process. Furthermore, to isolate the target object from unrelated regions, we adopt the local editing trick~\cite{mokady2023null-text} with a mask during the interval $[0, 0.8T]$. As CogVideoX-5b does not provide ideal masks, we instead utilize the method proposed in~\cite{zhou2022detecting} to generate the necessary masks. The proposed approach is implemented on a single NVIDIA LS20 GPU with 48GB of VRAM, achieving an average processing time of approximately 4 minutes per video, making it suitable for near-real-time applications.

\section{Different Functions of Attention Blocks}
\label{sec:b}
In this section, we delve deeper into the different functions of attention blocks in the T2V model. In Figure~\ref{fig: attn-supp}, we visualize the attention output and attention map for six blocks out of a total of 42. As the block number increases, the attention output contains progressively more high-frequency components, indicating that earlier blocks establish the low-frequency, foundational structure of the video, such as object placement and movement, while later blocks focus on high-frequency refinements and details. Furthermore, the attention in earlier blocks is densely concentrated along diagonal lines, while the attention in later blocks becomes more evenly distributed. This further demonstrates that earlier blocks emphasize the key shapes and correspondences in the video, while later blocks focus on fine details. For video editing tasks, which require maintaining low-frequency features like object location and general shape while modifying high-frequency features, the best performance is achieved by using only the first four blocks, \textit{i.e.}, the sketching blocks.

\section{More Discussions}
\label{sec:c}
\begin{figure}[h]
\vspace{-10pt}
    \centering
    \includegraphics[width=0.99\linewidth]{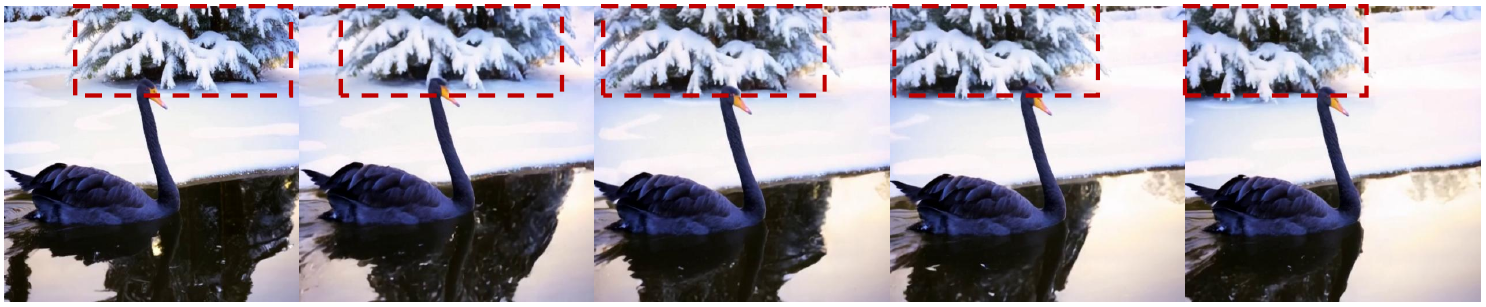}
    \caption{The generated video features a dynamic background, achieved by changing the prompt.}
    \label{fig: background}
    \vspace{-15pt}
\end{figure}

\textbf{About the static background.} The static background in the winter example is due to the limited prompt. As shown in Figure~\ref{fig: background}, by explicitly adding `\textit{moving forward}' to the prompt, we can generate a dynamically moving background.

\par \noindent \textbf{About the ablation results.} Using all blocks and the entire spectrum for guidance can slightly improve the preservation (M.PSNR, SSIM) since it leverages more video features for reconstruction. However, it introduces two major issues: (1) Editing quality (CLIP) may degrade, as features in sharpening blocks often need not remain unchanged, and (2) The additional blocks increase inference time and GPU memory usage (67.6GB for all blocks), making it less feasible. 
\par
\begin{figure}[h]
\vspace{-10pt}
    \centering
    \includegraphics[width=0.99\linewidth]{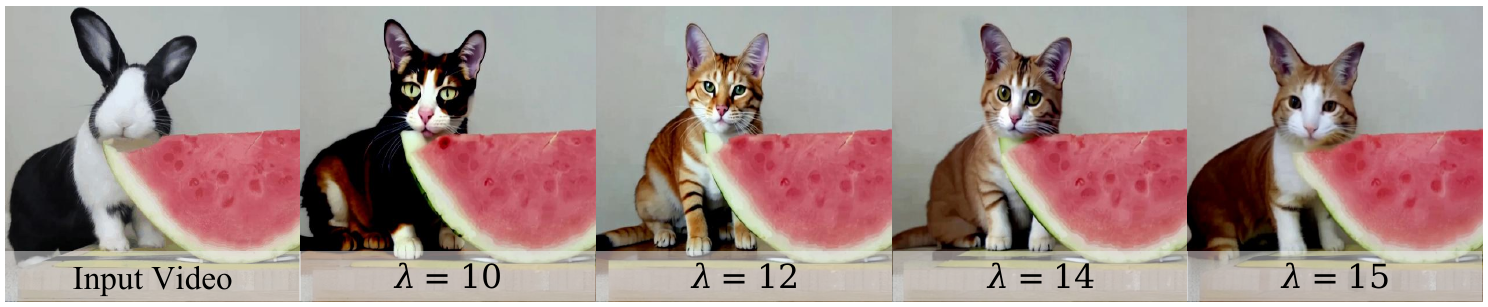}
    \caption{The sensitivity of $\lambda$.}
    \label{fig: lambda}
    \vspace{-5pt}
\end{figure}

 \noindent \textbf{About the sensitivity of $\lambda$.} Since we normalize the scale of the gradient in Eq.7, a reasonable range of $\lambda$ is [10, 15], which consistently produces plausible results. As illustrated in Figure~\ref{fig: lambda}, a larger $\lambda$ improves preservation but slightly reduces textual alignment. 

\par \noindent \textbf{About computational cost.} Our method is zero-shot and only involves sampling time (Table 2). With CPU offloading, our method uses 23.6GB GPU memory, comparable to Video-P2P (25.7GB), AnyV2V (17.1GB) and TokenFlow (11.4GB).  
\par \noindent \textbf{About the performance.} We acknowledge that our method’s quantitative improvement is not very significant compared with the existing methods, which heavily rely on \underline{\textit{one-shot tuning}} to enhance temporal consistency and adapt to the input sample. However, our approach is \underline{\textit{zero-shot}}. For a fair comparison, we found that fine-tuning our T2V model on the input sample significantly improves performance, as shown in Table~\ref{table:tuned}.



\begin{table}[h]
\centering
\caption{Quantitative comparisons under fine-tuning scenarios}
\vspace{-10pt}
\adjustbox{width=\linewidth}{\begin{tabular}{lccccc}
\toprule
Method  &
CLIP $\uparrow$&
M.PSNR $\uparrow$ & 
LPIPS $\downarrow$  &
OSV $\downarrow$ &
PF $\uparrow$ \\
\midrule
Tune-A-Video [\textcolor{cvprblue}{34}] &0.3522 &19.86 &0.4625 &35.01 & 0.09\\
Video-P2P [\textcolor{cvprblue}{15}] &0.3589 &20.57 &0.3199 &34.71 & 0.14\\
TokenFlow (ICLR'24)  &0.3614 &20.39 &0.3212 &34.50 & 0.12\\
\midrule

\rowcolor{gray!25}
FADE
 & 0.3762 & 20.69
 &0.3085& 31.36 & 0.27 \\
 \rowcolor{gray!25}
FADE (tuned)
 & \textbf{0.3946}& \textbf{22.19}
 &\textbf{0.2937} & \textbf{30.27} & \textbf{0.38} \\
\bottomrule
\end{tabular}
}

\label{table:tuned}
\vspace{-10pt}
\end{table}
\par
\begin{figure}[h]
    \vspace{-15pt}
    \centering
    
    \includegraphics[width=\linewidth]{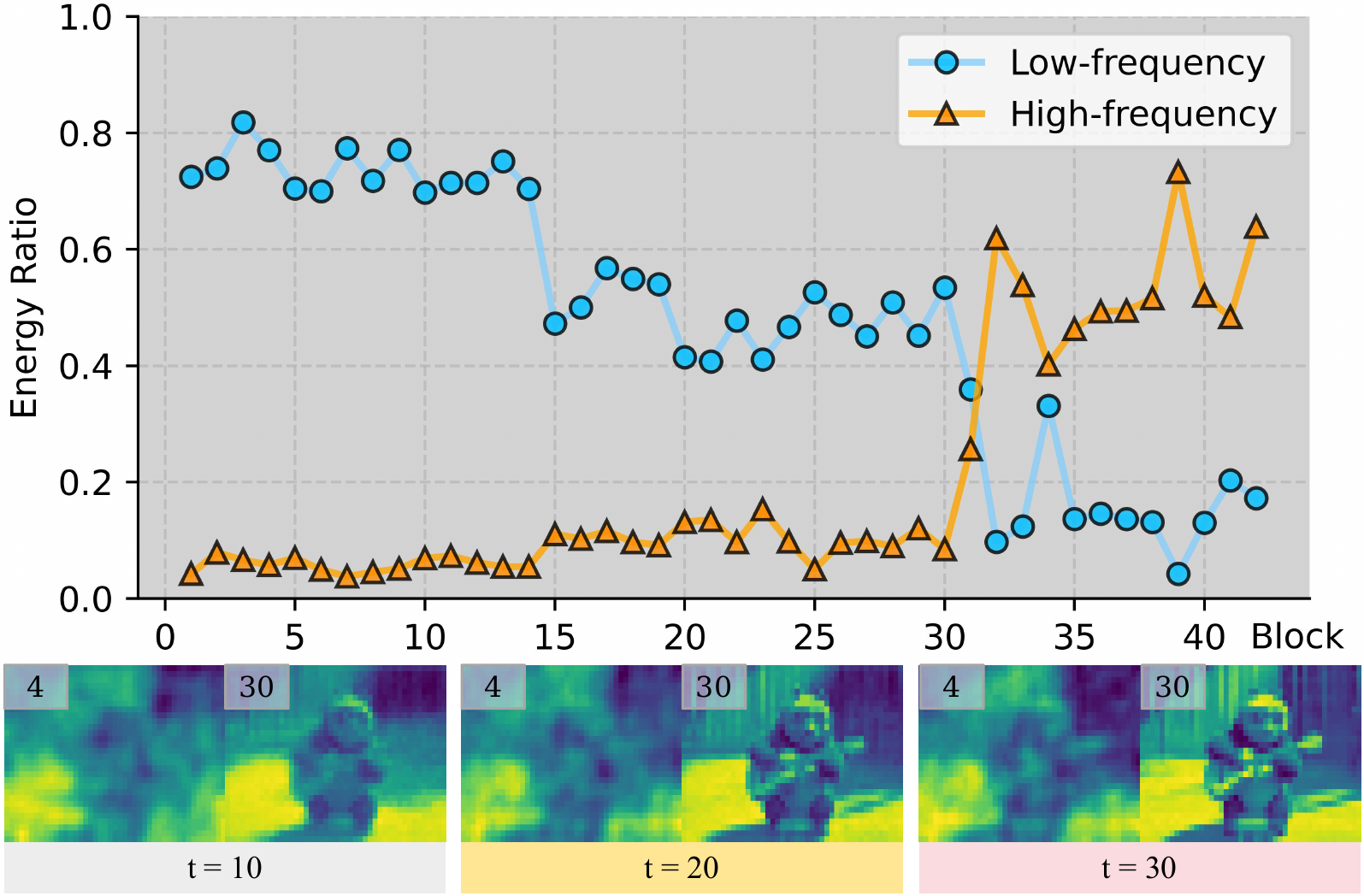}
    \caption{The spectral energy distribution and visualization of the attention results.}
    \label{fig: timestep}
    \vspace{-15pt}
\end{figure}

 \noindent \textbf{About video prior and block choice.} In Figure~\ref{fig: timestep}, We plotted the energy ratio of low and high-frequency components in the 3D DFT spectrum of the attention results at timestep=20 for all blocks. The frequency patterns align with the observations in Figure 2, supporting our motivation to distinguish between sketching and sharpening blocks. Moreover, this pattern remains consistent across sampling steps, as shown by the attention results in blocks [4] and [30] at timesteps $t=[10, 20, 30]$. 
    

\section{More Comparisons}
\label{sec:d}
\begin{figure}[h]
\vspace{-10pt}
    \centering
    \includegraphics[width=0.99\linewidth]{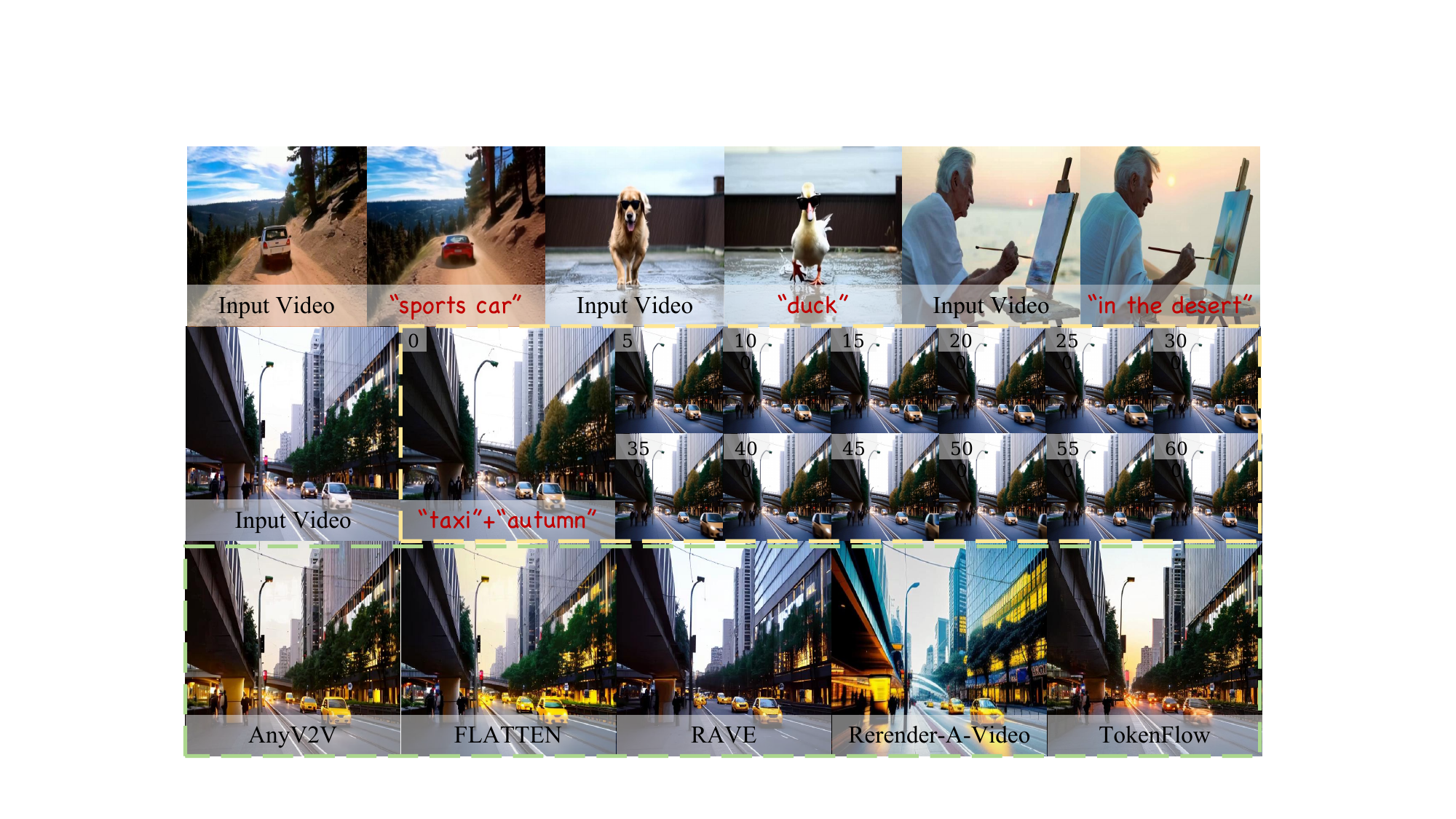}
    \caption{Qualitative results and comparisons}
    \label{fig: cmp}
    \vspace{-15pt}
\end{figure}

In this section, we add qualitative results (Row 1\&2) and comparisons with FLATTEN, TokenFlow, Rerender-a-Video, RAVE, and AnyV2V (Row 3), covering shape changes (sports car, duck), occlusion (desert, cat in the second figure of this PDF) and long video (taxi, $>$60 frames). Due to space limitations, we present one frame for comparison. As shown in Figure~\ref{fig: cmp}, our method achieves competitive performance and textual alignment.

\begin{figure*}[htp]
    \includegraphics[width=0.98\linewidth]{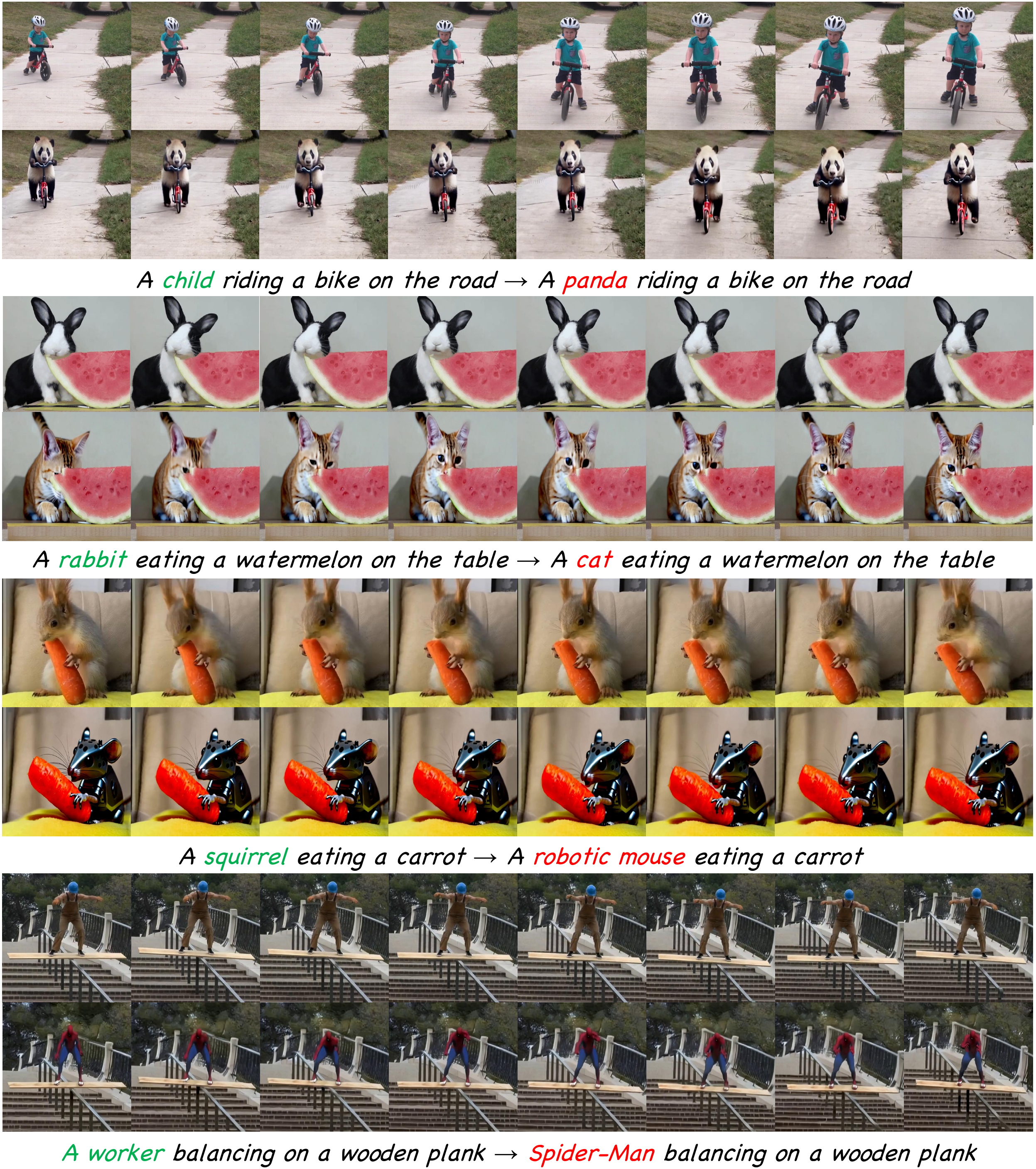}
    \caption{\textbf{More Qualitative Results.} FADE demonstrates impressive performance across a range of video editing tasks.}
    \label{fig: more-supp}
\end{figure*} 
\begin{figure*}[htp]
    \includegraphics[width=\linewidth]{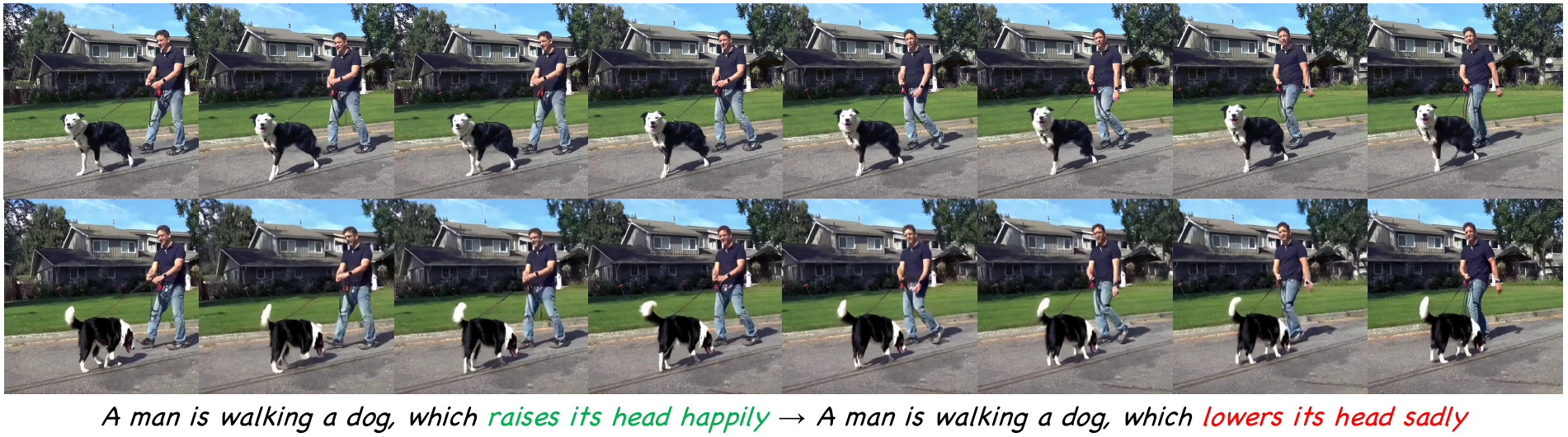}
    \caption{\textbf{Failure Case.} Our framework may occasionally generates artifacts, such as incorrect object orientation.}

    \label{fig: failure}
\end{figure*} 
\section{More Qualitative Results}
\label{sec:e}
We provide additional qualitative results to further demonstrate the effectiveness and versatility of our FADE framework across a variety of editing tasks. As shown in Figure~\ref{fig: more-supp}, our method produces highly realistic and coherent outcomes, even in challenging scenarios. For instance, in the first example, transforming “child” into “panda”, FADE successfully handles complex motions, such as riding a bike, which requires accurately modeling significant object movements from far to close perspectives. This highlights the framework’s ability to maintain spatial consistency during large-scale transformations. Another example, the transformation from “squirrel” to “robotic mouse”, emphasizes FADE’s flexibility in editing both textures and shapes. The results demonstrate the framework’s capacity to adaptively adjust fine details while preserving overall coherence, enabling seamless and visually plausible edits.

\section{Failure Cases}
\label{sec:f}
While our framework achieves impressive results across various editing tasks, it still faces limitations in handling certain complex scenarios. As illustrated in Figure~\ref{fig: failure}, our method occasionally produces artifacts, such as incorrect object orientation compared to the source video. For example, the dog that should be facing left in the edited output incorrectly faces right. This issue arises from the absence of specific constraints to guide object placement, resulting in random variations in the generated object’s orientation. Moreover, challenges also emerge when the edited objects are partially or fully occluded.

\end{document}